\documentclass[journal]{IEEEtran}
\usepackage[cmex10]{amsmath}
\usepackage[cmex10]{amsmath}

\usepackage{ifpdf}

\usepackage{algorithm}
\usepackage{algpseudocode}
\usepackage{xcolor}
\usepackage{url}
\ifpdf
\usepackage[pdftex]{graphicx}
\else
\usepackage[dvips]{graphicx}
\fi
\usepackage{tabularx}
\usepackage{multirow}
\graphicspath{{graphics/png/}}
\DeclareGraphicsExtensions{.png, .eps, .pdf}
\usepackage{float} 
\usepackage{subfigure}
\usepackage{comment}
\usepackage{bm}
\usepackage{amsmath,amsthm,amsfonts,amssymb}
\usepackage[noadjust]{cite}
\usepackage{nomencl}
\usepackage{hyperref}

\theoremstyle{definition}

\newcommand\Tstrut{\rule{0pt}{2.6ex}}         

\IEEEoverridecommandlockouts
\begin{document}
%
\title{Transferable Energy Storage Bidder}

\author{
    Yousuf Baker,
    \and Ningkun Zheng,~\IEEEmembership{Graduate Student Member,~IEEE},
    \and Bolun Xu,~\IEEEmembership{Member,~IEEE}
	\thanks{Y.~Baker, N.~Zheng, and B.~Xu are with Columbia University, NY, USA (e-mail: \{ykb2105, nz2343,  bx2177\}@columbia.edu). Y.~Baker was supported by the UAE Ministry of Education, N.~Zheng and B.~Xu were supported in part by National Science Foundation under grant ECCS-2239046.}
	}



\maketitle

\begin{abstract}
Energy storage resources must consider both price uncertainties and their physical operating characteristics when participating in wholesale electricity markets. This is a challenging problem as electricity prices are highly volatile, and energy storage has efficiency losses, power, and energy constraints. This paper presents a novel, versatile, and transferable approach combining model-based optimization with a convolutional long short-term memory network for energy storage to respond to or bid into wholesale electricity markets. 
We test our proposed approach using historical prices from New York State, showing it achieves state-of-the-art results, achieving between 70\% to near 90\% profit ratio compared to perfect foresight cases, in both price response and wholesale market bidding setting with various energy storage durations. We also test a transfer learning approach by pre-training the bidding model using New York data and applying it to arbitrage in Queensland, Australia. The result shows transfer learning achieves exceptional arbitrage profitability with  as little as three days of local training data, demonstrating its significant advantage over training from scratch in scenarios with very limited data availability.

\end{abstract}


\IEEEpeerreviewmaketitle

\section{Introduction}

Successful participation of energy storage resources in competitive electricity markets benefits storage investors and social welfare. Ancillary services such as frequency regulation have been the primary sources of revnue for energy storage owners, but these markets have quickly saturated due to surging storage deployments and small market size~\cite{mcgrath_comstock_2022}. In the meantime, the share of storage arbitraging in wholesale markets has tripled from a little less than 20\% in 2016 to almost 60\% in 2021~\cite{mcgrath_comstock_2022}. Thus price arbitrage in wholesale markets will be the main focus for future grid-scale energy storage projects.

Energy storage arbitrages price differences and earns revenues in wholesale energy markets, i.e., charging during low-price periods and discharging during high-price periods. At the same time, arbitrage from energy storage helps reduce renewable curtailments, meet peak demands, mitigate extreme events, and reduce the cost of electricity~\cite{SRIANANDARAJAH2022113052, article}. As countries and regions ramp up decarbonization efforts, energy storage resources are taking on an increasingly important role in future electricity markets and are becoming a cornerstone for cost-effective decarbonization~\cite{room2021fact, climate}. Thus, both energy storage owners and market organizers have significant economic and welfare drivers to evolve models and algorithms for energy storage to arbitrage robustly and profitably.

However, energy storage arbitrage is non-trivial due to highly volatile electricity prices and limited storage capacity. Various methods have been proposed in the literature to address energy storage participation in wholesale markets based on different theories. They require  dedicated  location-specific tuning and excessive computing power to achieve competitive arbitrage performance~\cite{sioshansi2021energy}. This paper proposes a novel end-to-end system for opportunity value calculation, prediction, and control, combining model-based dynamic programming with neural networks. Our approach innovates and provides several advantages as follows:
\begin{itemize}
    \item Our approach innovatively predicts the derivative of value-to-go functions, which represent the opportunity value of the state of charge, and uses dynamic programming to generate the training dataset. Compared to real-time prices, opportunity value functions are more stable and structured, thereby enabling easier prediction and contributing to the reliable performance of our approach;
    \item Our approach is highly computationally efficient in data pre-processing, training, and control. The complete training time, including generating training value functions, is less than six minutes over two years of price data, and operation/bidding decisions are generated instantly;
    \item Our approach utilizes transfer learning to maintain competitive performance over different markets and participation scenarios, including price response and market economic bidding;
    \item Our approach achieves state-of-the-art performance, achieving 70\% to near 90\% profit ratio compared to perfect foresight with various storage durations when tested using price data from New York, US, and Queensland, Australia.
\end{itemize}

The rest of the paper is organized as follows: Section~\ref{sec:lr} summarizes energy storage market participation and previous work using the learning method, Section~\ref{sec:form} and \ref{sec:solu} elaborates on the arbitrage formulation and solution method, Section~\ref{sec:result} presents the case study for price response and economic bid market rules in New York and the application of transfer learning for Queensland, and Section~\ref{sec:conc} concludes the paper.

\section{Literature Review}\label{sec:lr}

\subsection{Energy Storage Price Response and Self-Schedule}

Energy storage price response assumes the storage participant can observe the real-time price realization first and then decide on the operation privately without informing the system operator. The price response participation option primarily applies to small-scale behind-the-meter (BTM) storage resources ($<$ 1~MW)~\cite{roozbehani2012volatility}. Plenty of prior works have investigated energy storage price response using a variety of methods, including model-predictive control (MPC)~\cite{abdulla2016optimal}, stochastic programming~\cite{krishnamurthy2017energy}, approximate dynamic programming~\cite{jiang2015optimal}, and reinforcement learning~\cite{wang2018energy}. Price response is comparably an easier problem than economic bids as the storage operator is not limited to market clearing models and can act after observing new price signals. However, since price response mostly applies to small BTM storage projects, the revenue generated from arbitrage will unlikely justify any specialized computing hardware investments. Hence the arbitrage algorithm must be  slim and efficient to minimize the computation cost.

Alternatively, some markets allow energy storage operators to self-schedule and submit the operational schedule to the market operator. Still, this option is less frequently used in practice compared to participating by economic bids~\cite{7038219}. Self-scheduled storage cannot update the operation based on the system clearing price, a key difference compared to price-response or economic bids, which often causes the storage to miss price spike opportunities and deliver fewer market profits.


\subsection{Energy Storage Economic Bids}
FERC Order 841, issued in 2018, ordered all system operators in the US must allow storage to submit bids and cleared in spot markets~\cite{9776576}.
System operators allow storage participants  to submit charge and discharge bids at a specific time period ahead of the market clearing, usually one hour (also called hour-ahead bidding). The storage participant must follow market clearing results to charge or discharge, unlike in price response cases in which the storage can privately decide the control decision after observing the price. 

The bid design adds another layer of complexity in arbitraging, as optimal bid design requires mathematical tools due to storage SoC constraints. Wang et al.~\cite{wang2017look} formulate the energy storage look-ahead profit maximization problem as a bi-level optimization problem. A second approach for energy storage arbitrage control is backward dynamic programming~\cite{puterman2014markov}, and then the evolution is approximate-dynamic programming. Jiang and Powell outline a general approximate-dynamic programming framework for policy generation for energy storage operating with a stochastic generation source in response to stochastic demand~\cite{https://doi.org/10.48550/arxiv.1401.1590}, and further introduce a ``distribution-free'' variant of the previous algorithm that does not make any assumption on the price process~\cite{jiang2015optimal}. However, all of these methods are held back by large computational costs that make them hard to implement in real-world applications of arbitrage. 

There are other algorithms for energy storage real-time arbitrage control: Wang and Zhang~\cite{DBLP:journals/corr/abs-1711-03127} solve the arbitrage problem using reinforcement learning to come to an optimal arbitrage policy, and Zheng et al.~\cite{9721005} outline a computationally efficient analytical stochastic dynamic programming algorithm (SDP) for the problem of real-time price arbitrage of energy storage. Krishnamurthy et al.~\cite{7892020} also propose an SDP algorithm for arbitrage under day-ahead and real-time price uncertainties. However, none of the methods outlined above demonstrate or address transferability between different ISO zones and geographic locations, or the hour-ahead bid submission requirements in most real-time markets. 

\subsection{Machine Learning for Storage Arbitrage}
Recent efforts to apply machine learning for storage arbitrage can be grouped into two thrusts: the first is to use machine learning to generate price predictions and the second is to integrate them with MPC. In this case, the learning module is independent of the storage model.
Sarafraz et al.~\cite{5874828} and Nwulu and Fahrioglu~\cite{nwulu_fahrioglu_2013} outline two machine learning approaches for predicting locational marginal price (LMP) prediction using neuro-fuzzy logic and soft computing respectively, and Chaweewat and Singh~\cite{RNNPRICE} propose a residual neural network approach to price interval prediction. The main difficulty in combining price prediction with storage optimization is storage arbitrage requires a look-ahead of at least 24 hours to capture the daily price cycles~\cite{abdulla2016optimal}, while most real-time prediction methods may only accurately generate a few steps ahead of time. To this end, existing MPC approaches rely on pre-scheduling storage using day-ahead prices but have to neglect the real-time price variability, which is significantly higher than in day-ahead prices~\cite{krishnamurthy2017energy}. Our proposed method differs from MPC because it predicts the opportunity value function in the energy storage arbitrage problem. Our approach solves a single-period optimization problem based on the predicted value function, while MPC must solve a multi-period optimization. MPC is also sensitive to the volatility in the price prediction over a look-ahead horizon of at least 24 hours due to the arbitrage nature, while our method is immune to instabilities in long-horizon predictions as we only predict the value function over a single period.

The second approach is to directly use machine learning, mainly reinforcement learning (RL), to learn the optimal control policy for storage arbitrage directly. Wang et al.~\cite{wang2018energy} developed the first RL approach to arbitrage storage in real-time markets.
Cao et al.~\cite{9061038} propose a deep reinforcement learning approach to learn an optimal control policy for energy storage arbitrage with consideration of battery degradation. Kwon et al.~\cite{kwon2022reinforcement} demonstrated RL could optimize more sophisticated storage models in arbitrage by integrating battery degradation into the model.
Yet, a common disadvantage of RL-based approaches is transferability, as the model must undergo time-consuming training to be adapted to a new price zone or market environment. Transferability is a crucial aspect of storage arbitrage due to spatial and temporal variations: a typical system consists of hundreds of price nodes, and system price behaviors evolve with changes in system resource mix and ambient climate conditions. While previous efforts have looked into combining transfer learning with RL~\cite{taylor2009transfer} and its application in selected energy-related  issues, including demand response prediction~\cite{PEIRELINCK2022100126}, event identification~\cite{li2022transfer}, and battery health forecast~\cite{KIM2021102893}. Yet, the transferability of the storage arbitrage model has not been studied previously.


\section{Problem Statement and System Outline}\label{sec:form}
Our algorithm aims to predict the opportunity value at the current state of charge (SoC) of energy storage to maximize the price arbitrage profit. Our system consists three components: \emph{valuation}, \emph{forecasting}, and \emph{arbitrage}. We will first present our methods for valuation and arbitrage and then combine them with our forecasting model to form our bidding algorithm. We define $Q_t(e)$ as the opportunity value function representing the monetary value of the SoC $e$ at time step $t$. The problem formulation is adapted from ~\cite{xu2020operational, zheng2022comparing}, in which the solution is formulated using dynamic programming as follows:
\begin{subequations}\label{eq:p1}
\begin{align}
    \max_{ \substack{b_t, p_t, e_t \\ \in \mathcal{E}(e_{t-1})}} \lambda_t (p_t-b_t) - cp_t + \hat{Q}\big(e_{t}|\bm{\theta}, \mathbf{X}) 
    \label{eq:obj1}
\end{align}
where the first term is arbitrage revenue which is the product of the real-time market price $\lambda_t$ and the energy storage dispatch decision $(p_t-b_t)$, where $p_t$ is the discharge power and $b_t$ is the charge power. The second term is the discharge cost, where $c$ is the marginal operational cost, including degradation cost. The third term $\hat{Q}$ is the predicted storage opportunity value function with respect to SoC $e_t$.  The dynamic programming approach evaluates the energy storage by back-propagation, which is not viable in the real-time market where we do not have price realization ahead of time. Thus, we need to directly predict the value function $\hat{Q}$ using historical (and current) price data. $\hat{Q}$ is dependent on the prediction model parameters $\bm{\theta}$ and the prediction input features $\mathbf{X}$ over a look-back period.

We denote that the storage charge and discharge power and the final storage SoC belong to a feasibility set $\mathcal{E}(e_{t-1})$ which is dependent on the storage starting SoC $e_{t-1}$ at the start of time period $t$ (same as by the end of time period $t-1$). $\mathcal{E}(e_{t-1})$ is described with the following constraints:
\begin{gather}
    0 \leq b_t \leq P,\; 0\leq p_t \leq P \label{p1_c2} \\
    \text{$p_t = 0$ if $\lambda_t < 0$} \label{p1_c5}\\
    e_t - e_{t-1} = -p_t/\eta + b_t\eta \label{p1_c1}\\
    0 \leq e_t \leq E \label{p1_c3}
\end{gather}
\end{subequations}
where \eqref{p1_c2} models the upper bound, $P$, and lower bound, 0, constraints on the storage charge and discharge power. \eqref{p1_c5} is a relaxed form of the constraint that enforces the energy storage not charging and discharging simultaneously. Negative price is the necessary condition for storage to charge and discharge simultaneously in price arbitrage, hence by enforcing the storage to not discharge when the price is negative we eliminate simultaneous charging and discharging~\cite{xu2020operational}. \eqref{p1_c1} models the energy storage SoC evolution constraint with efficiency $\eta$ and \eqref{p1_c3} models the upper bound $E$ and lower bound (we assume as 0) of the storage SoC.


Creating our proposed system amounts to solving the problem of optimizing the prediction model parameters $\bm{\theta}$ to maximize storage arbitrage profit over a set of training price data and physical storage parameters. Intuitively, this problem can be formulated as a bi-level problem in which the upper level maximizes the total profit over the entire training time horizon. At the same time, the lower-level enforces a non-anticipatory decision-making process in which the storage dispatch decision only depends on the current price and the predicted value function as in \eqref{eq:p1}. However, this problem quickly becomes computationally intractable since the prediction model is embedded in the lower-level problem, formulated as a constrained optimization problem. Therefore, strong duality is required to convert the bi-level problem into a single-level equivalent problem or to derive partial derivatives and calculate the back-propagation gradients for gradient-based approaches. However, gradient-based approaches are complicated by the inclusion of SoC constraints~\cite{cui2017bilevel}.  In either case, the computational complexity quickly becomes overwhelming as the lower-level can include thousands of problems representing the arbitrage over a particular price data point. 

\textbf{Problem Statement.} We consider an alternative two-stage training approach in which we first generate the optimal opportunity value function and then train the learning model to predict the generated value function. These two stages are

\begin{itemize}
\item Opportunity value function generation:
\begin{subequations}\label{eq3}
\begin{align}
    Q_{t-1}(e_{t-1}) &= \max_{\substack{b_t, p_t, e_t \\ \in \mathcal{E}(e_{t-1})}} \lambda_t (p_t-b_t) - cp_t + Q_{t}(e_{t})     \label{eq:obj3}\\
    q_{t}(e) &= \frac{\partial}{\partial e}Q_t(e)
\end{align}
\item Neural network training:
\begin{align}
\min_{\theta} \sum_{e\in\mathcal{S}}\Big|\Big|\hat{q}_{t}\big(e|\bm{\theta}, \mathbf{X}) -  q_{t}(e) \Big|\Big|^2_2\label{eq:obj4}
\end{align}
\end{subequations}
\end{itemize}

Note that \eqref{eq:obj3} is also subject to the storage operation constraint set $\mathcal{E}(e_{t-1})$ as described in \eqref{p1_c2}--\eqref{p1_c3}. \eqref{eq:obj3} is a dynamic programming energy storage price arbitrage formulation in which the storage opportunity value is defined recursively as the maximized storage arbitrage profit including the profit from the current time step and the future opportunity values. This formulation fits a piece-wise linear approximation of the value function $q_t(e)$ based on the first order derivative of the optimal value function $Q_t$, and $e$ is from the set of SoC segments $\mathcal{S}$.  Note that in this formulation the prediction model parameters $\bm{\theta}$ are not involved in \eqref{eq:obj3}, hence this is a two-stage model in which we solve \eqref{eq:obj3} first and obtain all optimal value function results from $Q_t$, and more specifically, their derivatives $q_t$. We are then able to use \eqref{eq:obj4} to train the prediction model with the historical optimal value function at each time step.

\section{Solution and System Setup}\label{sec:solu}

\begin{figure*}[t]
    \centering
    \includegraphics[trim = 0mm 0mm 0mm 0mm, clip, width = 1.95\columnwidth]{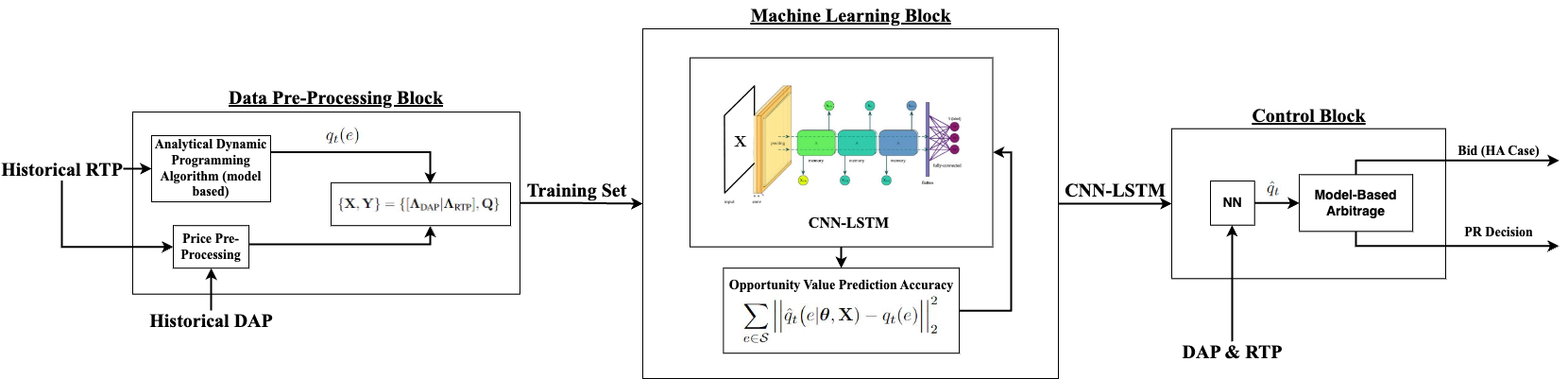}
    \caption{The proposed structure of training opportunity value function prediction model. }
    \label{fig:flow}
\end{figure*}

Our approach includes three steps: first, we use the deterministic price arbitrage dynamic programming approach to generate the optimal storage opportunity value function segments using historical price data. We then train a learning model to predict the optimal storage opportunity value segments using past price data. Finally, we test the learned model over unseen (future) price datasets. The system structure is shown in Fig.~\ref{fig:flow}, which includes the dynamic programming solution and the training method.

\subsection{Feature and Label Formatting \label{dataprep}}

In general, the spot price for energy exhibits long-term and short-term cycles according to cycling demand: the daily cycling between peak and non-peak hours and the long-term seasonal cycles; though events and the stochastic nature of price create differences in between. Thus we chose to use a convolutional long short-term memory (ConvLSTM) neural network, which can learn patterns in time series data. For learning timestep $t$, our network input/target pair could be $[ \lambda_t, q_t ]$ (or $q_{t+hr}$, where $+hr$ represents an hour time shift for the HA case). However, to better capture daily cycling, we elaborate our single-step input-output pair by constructing the following input-output matrices:
$$
\{ \mathbf{X}, \mathbf{Y} \} = \{ [\mathbf{\Lambda_{\text{DAP}}} | \mathbf{\Lambda_{\text{RTP}}}], \mathbf{Q} \}
$$
$$
\mathbf{\Lambda_{\text{DAP}}} = 
\begin{bmatrix}
\lambda_{\text{DAP}, t-m} & \lambda_{\text{DAP}, t- m + 1} & \ldots & \lambda_{\text{DAP}, t} \\
\lambda_{\text{DAP}, t-m-1} & \lambda_{\text{DAP}, t-m} & \ldots & \lambda_{\text{DAP}, t-1} \\
\vdots & \vdots & & \vdots \\
\lambda_{\text{DAP}, t- m - 5hr} & \lambda_{\text{DAP}, t-m-5hr+1} & \ldots & \lambda_{\text{DAP}, t-5hr}
\end{bmatrix}
$$
$$
\mathbf{\Lambda_{\text{RTP}}} = 
\begin{bmatrix}
\lambda_{\text{RTP}, t-n} & \lambda_{\text{RTP}, t-n + 1} & \ldots & \lambda_{\text{RTP}, t}\\
\lambda_{\text{RTP}, t-n-1} & \lambda_{\text{RTP}, t-n} & \ldots & \lambda_{\text{RTP}, t-1} \\
\vdots & \vdots & & \vdots \\
\lambda_{\text{RTP}, t-n-5hr} & \lambda_{\text{RTP}, t-n-5hr+1} & \ldots & \lambda_{\text{RTP}, t-5hr}
\end{bmatrix}
$$
$$
\mathbf{Q} = \begin{bmatrix}
q_t \\
q_{t-1}\\
\vdots \\
q_{t-5hr}
\end{bmatrix}
$$
where $\mathbf{\Lambda_{\text{DAP}}}$ and $\mathbf{\Lambda_{\text{RTP}}}$ are matrices made up of our day-ahead and real-time price data, and $m,n$ are a lookback window for the day ahead and real-time prices respectively, and $5hr$ is the number of timesteps that make up five hours in a given market resolution (60 in a 5 min resolution market). This allows the network to capture not only the information on past prices for the current value function but also the relationship between past value functions in a 5-hour lookback. We chose five hours here as it is long enough to capture cycles within a single day (e.g. peak vs non-peak demand and the transition between them). The inclusion of the day ahead price here serves as a more stable price reference for the corresponding hour's spot price. Also of note is the cyclic symmetry of the price matrices along the diagonal, which allows the network to learn better the equivariant properties of the dataset~\cite{cyclicsym}. Finally, the choice of a ConvLSTM, as opposed to a traditional LSTM, is to allow the network to capture the "vertical" temporal relation between the five hours of data in each data block. 

Note that for DAP, the shift applied to t across rows corresponds to a step shift in the resolution of the real-time market. Meaning that if it is a 5-minute resolution real-time market, the first 12 rows of $\mathbf{\Lambda_{\text{DAP}}}$ will be the same since the day-ahead market  is hourly resolution.

\subsection{Model Selection and Transfer Learning}

The focus of this paper is to demonstrate the robustness of the approach across different market conditions and battery durations and to show its transferability between zones. Thus we chose one general network architecture for testing. However, initial experimentation showed minimal gain-loss in network performance on minor parameter changes across cases. Further, to guarantee that the training converges to a well-performing set of weights, multiple networks were trained for each case. The weights achieving the most consistent and low validation error were saved for evaluation. Of those, the best model was chosen by the highest arbitrage profit. The network is trained over 100 epochs in the case where it is trained from scratch, and 25 epochs for the transfer learning training, with a learning rate of $10^{-3}$. Further, we use a callback function that saves the model weights only when the validation error improves, ensuring that the weights loaded for training are not overfitted. This callback also allows us to set our epochs with significant overhead to ensure convergence without over-fitting in all cases. 

Furthermore, we apply transfer learning to quickly adapt a trained model from one price zone to another. Our transfer learning approach freezes all model layers except the output layer and retraining on the dataset of the task to be transferred to~\cite{chollet2015keras}. The underlying assumption is that the output layer is more sensitive to data variability while the rest of the network captures persistent patterns in the data. Note that this transfer learning approach does not directly apply to storage of different durations as the number of outputs in the output layer depends on the storage duration.

\subsection{Full Algorithm}

We lay out our workflow below, which is a sequence of three algorithms. As a prerequisite for model training, we generate all value functions $\mathbf{Q}$ according to the dynamic programming solution in Appendix-\ref{dpapp}. After this, we construct our input dataset and train our prediction model according to Algorithm 1, which produces our trained model weights $\bm{\theta}$.

\begin{algorithm}
  \caption{Value Function Prediction Model Training}
  \begin{algorithmic}[1]
    \State \textbf{Dataset Preparation:} Pre-Process data according to \ref{dataprep}
    \State\textbf{Initialization:} Initialize model parameters $\bm{\theta}$ using random seed.
      \While{stop criteria not true}
        \For{$t \in [1,t]$}
            \State $\bm{x} \gets [\mathbf{\Lambda_{\text{DAP}}} | \mathbf{\Lambda_{\text{RTP}}}]$ 
            \State $\bm{y} \gets \mathbf{Q}$
            \State Calculate Loss Components by Eq.~(\ref{eq:obj4}) 
            \State \textbf{Update} $\bm{\theta}$ by backpropagation
        \EndFor
      \EndWhile
      \State \textbf{return} $\bm{\theta}$\Comment{Parameters of the prediction model} 
      
  \end{algorithmic}
  \label{alg:1}
\end{algorithm}

Algorithm 2 outlines a transfer learning approach to adapt the prediction model produced by Algorithm 1 for use by another zone. Algorithm 2 exhibits higher computational efficiency, as it follows a similar workflow as Algorithm 1, but with smaller training datasets specific to the new zone. The resulting trained model weights are denoted as $\bm{\theta^*}$. We differentiate between the two sets of model weights since we compare the two transferring approaches (transfer learning, and applying the model on new zones without retraining) later in the paper.

\begin{algorithm}
  \caption{Transfer Learning}
  \begin{algorithmic}[1]
    \State\textbf{Initialization:} Initialize model parameters $\bm{\theta^*}$ using random seed
    \State $\bm{\theta^*} \gets \bm{\theta}$ (trained model parameters)
    \State Freeze all parameters except output layer parameters 
    \State Repeat training loop using new region's data set $\{ [\mathbf{\Lambda^*_{\text{DAP}}} | \mathbf{\Lambda^*_{\text{RTP}}}], \mathbf{Q^*} \} $
    \State \textbf{return} $\bm{\theta^*}$\Comment{Parameters of the prediction model} 
      
  \end{algorithmic}
  \label{alg:2}
\end{algorithm}

Algorithm 3 outlines the process of simulating arbitrage using our prediction model in price response cases. The arbitrage simulation is as follows: use the prediction model trained in Algorithm 1 and/or Algorithm 2 to predict value functions using the current real-time price and a look-back window (including the day ahead look-back); then make the operational decision by solving the single-period optimization \eqref{eq:p1}. In the economic bids case, we predict value functions and then generate the bids instantaneously according to Appendix-\ref{bidgen}, as it simply down-samples the predicted value function modified by efficiency. Once the bids are generated, use them to simulate arbitrage and market clearing as outlined in Appendix-\ref{arbapp}.

\begin{algorithm}
  \caption{Arbitrage with Value Function Prediction}
  \begin{algorithmic}[1]
    \State\textbf{Initialization:}
    \State Set energy storage parameters $c, P, \eta, E $.
    \State Initialize $e_{t-1}\gets e_0$.
      \For{\texttt{$t\in [1,T]$}}
        \State Predict $\hat{q}\big(e_{t}|\bm{\theta}, \bm{X}\big)$
        \State Solve single-period optimization (\ref{eq:p1})
        \State \textbf{Return} $e_t, p_t, b_t$
      \EndFor
  \end{algorithmic}
\label{alg:3}
\end{algorithm}

\section{Case Study Set-ups}\label{sec:result}

\subsection{Market Participation Setting and Storage Parameters}
We consider the following four market designs and participation settings to demonstrate that our proposed approach fits a wide range of storage participation options and market designs:
\begin{itemize}
    \item \textbf{HA-1}. Energy storage owner submits \emph{single-segment} bids \emph{one hour ahead} to real-time markets. This represents the current storage bidding model in most wholesale real-time markets in the US~\cite{sakti2018review,jiang2015optimal} where energy storage submits one charge bid and one discharge bid one hour ahead of the market clearing. The storage can update its bid for each hour, but the bids must stay the same within each hour for multiple market clearings (for example, real-time markets clear every five minutes in NYISO, so one hour includes 12 real-time clearings).
    \item \textbf{HA-10}. Same to HA-1 except the storage submits \emph{10-segment} SoC-dependent charge and discharge bids. This is a new market design proposed for CAISO to economically manage storage SoC in real-time~\cite{zheng2022energy,chen2022convexifying}.
    \item \textbf{PR-10}. The storage conducts \emph{price response} in real-time, deciding the storage control after observing the published real-time price, instead of submitting bids~\cite{zheng2022arbitraging}. The price response option is limited to behind-the-meter storage in which the associated demand is cleared in real-time market prices. In this case, the storage is not limited to any bidding models and can use any decision-making models. Yet, we assume the storage uses a 10-segment approximation of its opportunity value as it provides a good enough approximation to the actual value function. This also enables us to benchmark HA-10 and PR-10 cases to demonstrate the economic cost of the hour-ahead bidding requirement.
    \item \textbf{PR-1}. Same as PR-10 except the storage uses the average opportunity value (i.e., \emph{one segment} approximation) for arbitrage control. This is not a realistic case as there is no motivation for the storage operator to limit itself to using a single-segment, less accurate approximation of its value function to conduct arbitrage. However, we include this case with the sole purpose to benchmark against the HA-1 case and PR-10 case.
\end{itemize}

In all case studies, we consider storage with a 90\% one-way efficiency and a \$10/MWh cost of discharge (excluding the opportunity cost), unless otherwise specified. We consider three storage durations including 2-hour, 4-hour, and 12-hour. Further, we adapt our base prediction model to predict the hour ahead case by adding an hour time shift to our ground truth training target value function, which corresponds to 12-time steps in 5-minute price resolution.

We conduct the majority of our case studies over price data from New York ISO (NYISO)~\cite{nyisodat} for four price zones: NYC (Zone J), LONGIL (Zone K), NORTH (Zone D), and WEST (Zone A). We also use data from the Australian Energy Market Operator (AEMO) for Queensland to demonstrate the transferability of our approach using transfer learning~\cite{uqdat}.


\subsection{Market and Price Data}
We observe differences in price statistics and generation mix across zones from the same ISO, and in between zones from ISO's in other states and even countries, summarized in Table \ref{tab:PriceStats}. In NYISO, these differences can be attributed to significant transmission congestion when comparing the two main zone groups in NYISO~\cite{patton20162014}.
QUEENSLAND has the highest price volatility, which can potentially be attributed to the absence of a day-ahead market. Further, we see a clear tie between penetration rates of renewables into the zones and the price volatility~\cite{waite2020electricity}. We also see the highest occurrence of negative prices in NORTH, which is due to the significantly higher penetration of wind when compared to NYC and LONGIL.
\begin{table}[ht]
\footnotesize
\caption{Price Data Statistics}, 
\centering{
\begin{tabular}{cccc}
\hline
\hline
\Tstrut
Zone & Negative Price \# & STD
& Renewable \%  \\
\hline
\text{NYC (NY)}&208&28.82&0.93\\
\hline
\text{LONGIL (NY)}&190&50.17&0.93\\
\hline
\text{NORTH (NY)}&6334&40.25&13.06\\
\hline
\text{WEST (NY)}&633&37.55&13.06\\
\hline
\text{QNSLND (AUS)}&522&243.00&13.19\\
\hline
\hline
\end{tabular}
}

\label{tab:PriceStats}
\end{table}

All codes for valuation, network training, and arbitrage are written in Python with Jupyter Notebook and is available on GitHub\footnote{\url{https://github.com/ybaker661/LSTM-Value-Prediction}}. All trials are run on a desktop computer with AMD Ryzen 9 processor and Nvidia GPU on Tensorflow 2.9.1 and with cuDNN and CUDA versions 8.1 and 11.2, respectively. All case studies using price data from NYISO were trained using data from 2017 to 2018, and tested over 2019 data. Each year of price data for each price zone has 8760 day-ahead price data points (hourly resolution) and 105,120 real-time price points (5-minute resolution). The look-back price window includes the last 36 real-time prices (3 hours) and 24-day-ahead prices (one day).
The \textit{maximum training time} over two years of training price data, including the generation of historical optimal value functions and training of the neural network, is \textit{390 seconds}, a bit more than five minutes. The network consists of a Convolutional Block with three sequential time-distributed Convolutions+MaxPool layers, then an LSTM block with two sets of bi-directional LSTM+drop out layers, and then finally a Dense layer at the output end. The specific model hyperparameters and details can be found on GitHub.

\section{Results}
\subsection{Benchmark with Competing Methods}
\begin{figure*}[ht]%
\centering
    \subfigure[Accumulated]{\includegraphics[trim = 46mm 0mm 40mm 0mm, clip, width = 2\columnwidth]{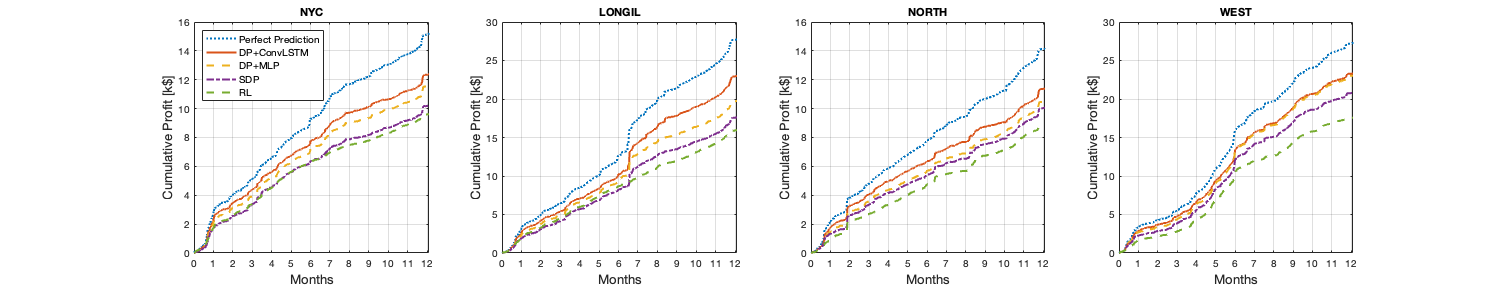}\label{f}}
    \hfill
    \subfigure[Monthly]{\includegraphics[trim = 45mm 0mm 42mm 0mm, clip, width = 2\columnwidth]{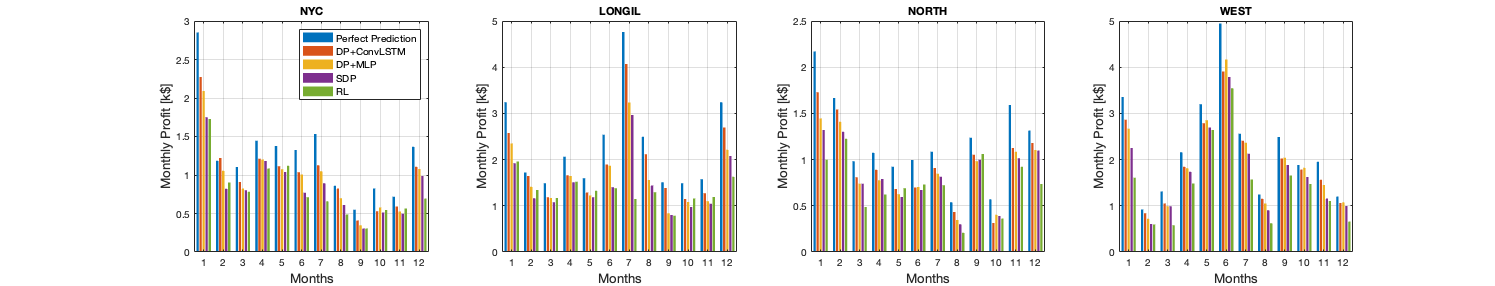}\label{fig.monthly}}
\caption{Profit over 2019 test set for NYISO Zones (a) Accumulated and (b) Monthly. The subfigures presented in the same row share a common legend. In (b), the order of columns within each bin conforms to the top-down ordering of categories presented in the legend.}\label{fig*:RLbench}
\end{figure*}

We first benchmark our proposed approach with other competing energy storage price arbitrage methods in a price response setting (PR-10), in which storage can observe price first and act accordingly, without bidding ahead into markets. We benchmark the proposed method (DP-ConvLSTM) with a reinforcement learning method (RL)~\cite{DBLP:journals/corr/abs-1711-03127}, a modified stochastic dynamic programming with day-ahead price updates (SDP)~\cite{zheng2022arbitraging}, the proposed method but implemented with a multilayer perceptron (DP-MLP) network~\cite{zheng2022energy2}, and perfect price predictions which provide the highest profit possible. In RL, we have 11 actions, 103 price states, and 121 SoC states, which take more than 1 hour to train for 5-min resolution arbitrage. The RL approach uses a Markov decision process (MDP) model by discretizing the storage SoC, and it only works with perfect efficiency (100\%). To provide a fair comparison, all methods in benchmark comparison consider storage with perfect efficiency, while other cases still consider 90\% efficiency.

Fig.~\ref{fig*:RLbench} shows the comparative results when trained using price data from 2017-2018 and tested in 2019 at price zones in NYISO. The result shows that DP-ConvLSTM has a clear advantage over other methods in terms of profitability. Notably, the DP-ConvLSTM approach performs exceptionally well in capturing low-frequency extreme events, such as the surge in profits around June in LONGIL and WEST, where the ConvLSTM captures profit spikes that the RL benchmark misses. The difference between the ConvLSTM profit value and the perfect prediction comes from the difference in arbitrage decision as a result of numerical saturation. In this context, numerical saturation means that the network learns to predict numerical values in the range of data it most frequently sees (value functions of stable prices), and so when it predicts on anomaly data (price spike value functions) that are numerically much larger, the network prediction saturates at the largest common numerical value it sees.

\subsection{Price Response}
\begin{table}[ht]
\footnotesize
\caption{Captured percentage profit ratios: Price-response}
\centering
\begin{tabular}{lcccccc}
\hline
\hline
\Tstrut
Zone   & \multicolumn{3}{c}{PR-1} & \multicolumn{3}{c}{PR-10} \\
 & 2hr         & 4hr       & 12hr & 2hr         & 4hr       & 12hr\\
\hline
NYC       &80.83 &79.96 &73.54 &83.69 &83.63 &75.67 \\
LONGIL    &82.33 &80.61 &79.10 &82.98 &83.94 &82.38 \\
NORTH     &78.24 &75.87 &71.02 &79.52 &79.96 &74.29 \\
WEST      &84.43 &80.37 &83.65 &87.97 &87.44 &84.43 \\
\hline
\hline
\end{tabular}
\label{tab:PriceResponse}
\end{table}
In this subsection, we compare the price response arbitrage performance (PR-1 and PR-10) with different storage durations. Table~\ref{tab:PriceResponse} shows the arbitrage profit ratio results.

Overall, the result shows stable performance over the four price zones and three storage durations. In comparison, our previous work using SDP~\cite{zheng2022arbitraging} and DP-MLP~\cite{zheng2022energy2} have worse performance in LONGIL (more frequent price spikes) and NORTH (more frequent negative prices). The comparison between PR-1 and PR-10 shows that increasing the value function approximation from one to ten segments increased the profit ratio by around 3\%. Considering different storage durations, the profit ratio is lower for long-duration energy storage (12hr), as the longer storage duration leads to a longer temporal correlation into the future, thus, higher prediction difficulties. Still, our method achieved around 75\% profit ratio (PR-10) in the worst-case scenario in the NORTH zone.

\subsection{Hour-ahead Bidding}
\begin{table}[ht]
\footnotesize
\caption{Captured percentage profit ratios: Hour-ahead}
\centering
\begin{tabular}{lcccccc}
\hline
\hline
\Tstrut
Zone   & \multicolumn{3}{c}{HA-1} & \multicolumn{3}{c}{HA-10} \\
 & 2hr         & 4hr       & 12hr & 2hr         & 4hr       & 12hr\\

\hline
NYC       &73.99 &74.82 &77.00 &78.79 &80.61 &74.47\\
LONGIL    &74.26 &76.56 &82.01 &75.30 &79.63 &81.89\\
NORTH     &73.22 &71.71 &70.17 &75.83 &77.21 &73.16\\
WEST      &78.79 &80.13 &84.17 &83.12 &83.94 &84.60\\
\hline
\hline
\end{tabular}
\label{tab:HANY}
\end{table}

We now investigate hour-ahead bidding, which is the most common market design for energy storage owners operating in the real-time market, where the storage submits bids an hour ahead of time. Table~\ref{tab:HANY} shows the hour-ahead bidding profit ratio in the NYISO case study. The profit ratio is lower than the price response as the storage owner must decide on the bids one hour before the actual time of arbitrage. The short-duration storage (2hr) cases have higher profit ratio reductions (up to 7\%) as the value function is more sensitive to recent market prices. On the other hand,  the long-duration storage (12hr) is more resilient and the hour-ahead bidding has little impact on the profit ratio.

Hour-ahead bidding results also restate our observation from the price response case, that multi-segment SoC bids are more beneficial for short-duration storage to better manage their SoC, while the improvement is not obvious for long-duration storage. Overall, our approach achieved a higher than 70\% profit ratio in all hour-ahead cases, showing robust performance under different market designs and storage technologies.

\subsection{Transfer Learning in AEMO}

\begin{table*}[t]
\footnotesize
\caption{Percentage profit ratio for hour-ahead bidding in QUEENSLAND AUS with different amounts of training data}
\centering
\begin{tabular}{llrrrrrrrrrr}
\hline
\hline
  &  & \multicolumn{5}{c}{HA-1}   & \multicolumn{5}{c}{HA-10}               \\
Duration   & Training & No Data & 3 Days & 1 Week & 1 Month & 1 Year & No Data & 3 Days & 1 Week & 1 Month & 1 Year  \\
\hline
\multirow{2}{*}{2hr}    & T.L.           & 77.24      & 82.76       & 82.85      & 81.30           & 85.35 & 78.90      & 84.00       & 81.54      & 79.93           & 83.42\\   
                        & No T.L.       & X     & 48.22       & 51.36      & 78.59        & 83.88 & X      & 44.59       & 44.88      & 78.10      & 86.95\\
\hline
\multirow{2}{*}{4hr}    & T.L.           & 81.21     & 81.29       & 81.31      & 78.11      & 79.12 &      84.06  &87.44      & 84.34      & 77.79      & 82.65\\   
                        & No T.L.       & X     & 62.40     & 65.45      & 74.92          & 80.42 & X     & 55.99      & 55.99      & 74.36      & 83.11\\
\hline
\multirow{2}{*}{12hr}    & T.L.            & 92.43     & 83.96       & 81.32      & 78.74      & 82.73 & 90.69    & 80.78       & 79.11      & 78.84            & 81.35\\   
                        & No T.L       &  X    & 74.79       & 75.53      & 74.39      & 75.43 & X    & 73.65       & 73.59      & 74.56    & 82.36\\

\hline
\hline
\end{tabular}
\label{tab:QNSHA}
\end{table*}

We now demonstrate the effectiveness of applying transfer learning to quickly adapt a pre-trained value function prediction model from one market to a new market. In this case study, we pre-train the prediction model using NYC price data from 2017-2018 and conduct arbitrage in Queensland, Australia. In Queensland, we use selected data from 2019 for training and the first 6 months of 2021 for evaluation. We skipped the year 2020 because of COVID-19's impact. 
To present the sensitivity of transfer learning over a limited amount of data, we consider various durations of training dataset. 
We present a sensitivity analysis comparing the performance of transfer learning versus training a model from scratch for the situations where we have access to training data for only 3 days, 1 week, 1 month, and 1 year of data for the target zone. 
Thus this case study has the following steps:
\begin{enumerate}
    \item Use a pre-trained network (transfer learning) or a randomly initialized network (training from scratch).
    \item Use a limited duration of Queensland price data from 2019, ranging from 3 days to 1 year, to train the model using transfer learning as outlined in algorithm 2, or normal training outlined in algorithm 1.
    \item Test the arbitrage performance to arbitrage, as outlined in algorithm 3, using the first six-month of data in Queensland, 2021.
\end{enumerate}

Table~\ref{tab:QNSHA} shows the arbitrage profit ratio results for Queensland. 
The transfer learning approach vastly outperforms training a model from scratch in data-scarce scenarios. Note that our approach with transfer learning outperforms the University of Queensland St Lucia Tesla Battery Project, which uses a commercial-grade model prediction control. Our approach achieves around 80\% profit ratio compared to their reported 68.2\% profit ratio for a 2-hour battery in the AEMO market~\cite{wilson_esterhuysen_haines}. We also see that adding more data to the transfer learning case does not necessarily increase performance, whereas training the model from scratch becomes a viable option once a certain amount of data is available. For the 2-hour storage, training from scratch becomes viable when you have one month of data available for the target zone. For the 4-hour storage, the model still needs about one month of data to reasonably perform when trained from scratch, though the model can capture higher profit ratios for three days and one week of data than the 2-hour case. Compared to both of these, the 12-hour storage seems to be the easiest for the model to learn, only needing three days of data when training from scratch to achieve reasonable performance; however, the 12-hour storage shows that the transfer learning approach outperforms training from scratch for all data scenarios for 1 and 10 segments. However, since the 12-hr storage has a lower opportunity cost and less significant change in the opportunity cost between sequential time steps, predicting the opportunity value might not be an effective method.

The takeaway is that transfer learning beats training the model from scratch when data scarcity is an issue. However, when the dataset size increases to a general size of 1 month, training from scratch becomes a viable option. Additionally, adding extra data, past three days or one week for transfer learning and the past~6 months for training from scratch, does not necessarily yield better performance. As such, it is more useful to focus on stabilizing ConvLSTM's volatile and initialization-sensitive training and other changes to the training process. In almost all cases, using the model trained on NYC data without any retraining performs comparably or even better than transfer learning and even training a model from scratch. This indicates statistical robustness and generality in the NYC data, and it also points to a unified generating distribution behind the price data. However, further analysis is required regarding other permutations of transfer learning and on testing different data duration permutations.

\section{Conclusion}\label{sec:conc}
In this paper, we propose a computation-efficient, versatile, and transferable energy storage arbitrage model that fits both price response and market bidding. Our proposed approach achieves state-of-the-art profits compared to other methods and is both computation and data-efficient. We also demonstrate that by incorporating transfer learning, we can quickly adapt our bidding model to a new location with very limited training data. Our model suits a variety of arbitrage settings, including behind-the-meter price response and economic bids for utility-scale storage, and can be implemented using non-proprietary software and regular computing hardware. Our work would facilitate storage participation in electricity markets and promote economic decarbonization of the electric power system.

\bibliographystyle{IEEEtran}	
\bibliography{IEEEabrv,main}		

\appendix

\subsection{Dynamic Programming Solution Algorithm \label{dpapp}}

We first solve the dynamic programming problem as listed in \eqref{eq:obj3} subject to constraints \eqref{p1_c2}--\eqref{p1_c3}. We use results from our prior work~\cite{xu2020operational} to solve the dynamic programming problem \eqref{eq:obj3} and obtain the full piece-wise linear approximation of the opportunity value function $Q_t$ for all time periods (i.e., one value function for each time step for an entire year, 105120 for 5 min price resolution 35040 for 20 min price resolution). 
We start by defining $q_t$ as the derivative of storage opportunity value function $Q_t$, which represents the marginal opportunity value of energy stored in the storage. Then we can use an analytical formulation to calculate the opportunity value $q_t(e)$ at any given energy storage SoC level.

Our prior work proved $q_{t-1}$ can be recursively calculated with next period value function $q_t$, power rating $P$, and efficiency $\eta$. The value function calculated using the deterministic formulation is thus
\begin{align}\label{eq3-1}
    &q_{t-1}(e) = \nonumber\\
    &\begin{cases}
    q_{t}(e+P\eta)  & \text{if $\lambda_{t}\leq q_{t}(e+P\eta)\eta$} \\
    \lambda_{t}/\eta  & \text{if $ q_{t}(e+P\eta)\eta < \lambda_{t} \leq q_{t}(e)\eta$} \\
    q_{t}(e) & \text{if $ q_{t}(e)\eta < \lambda_{t} \leq [q_{t}(e)/\eta + c]^+$} \\
    (\lambda_{t}-c)\eta & \text{if $ [q_{t}(e)/\eta + c]^+ < \lambda_{t}$} \\
    & \quad\text{$ \leq [q_{t}(e-P/\eta)/\eta + c]^+$} \\
    q_{t}(e-P/\eta) & \text{if $\lambda_{t} > [q_{t}(e-P/\eta)/\eta + c]^+$} 
    \end{cases}
\end{align}
and calculates the opportunity value function assuming the price follows a recursive computation framework. This deterministic formulation is what we will use in our investigation, and from this we are able to calculate opportunity value function $q_t(e)$ at any time step using backwards recursion by defining an end period value function $q_T$. We then discretize $q_t$ by splitting the energy storage SoC level $e$ into small equally spaced segments, which must be far smaller than power rating $P$. For any SoC level $e_t$, we can find the nearest segment and return the corresponding value.

\subsection{Bid generation \label{bidgen}}

We now design discharge and charge bids using the opportunity valuation results based on our prior work~\cite{zheng2022comparing,zheng2022energy}. We consider generating time-varying SoC-dependent bids with a total number of $J$ segments for charge bids $B_{t,j}$ and discharge bids $C_{t,j}$. Note that these bids represent the combination of the discharge cost and the change in the opportunity value. We assume each bid segment $j$ is associated with an SoC range $E_{j-1}$ to $E_j$. The discharge bids are thus calculated based on the average value function between the internal $E_{j-1}$ and $E_j$
\begin{align}
    C_{t,j} &= \frac{1}{E} \int_{E_{j-1}}^{E_j} \frac{\partial}{\partial p_t} (c p_t - Q_t(e_{t-1} - p_t/\eta + b_t\eta)) d{e_{t-1}}\nonumber\\
    &= c+ \frac{1}{E} \int_{E_{j-1}}^{E_j} q_t(e_{t-1} - p_t/\eta + b_t\eta) d{e_{t-1}} /\eta \nonumber\\
    &\approx c+ \frac{1}{\eta E} \int_{E_{j-1}}^{E_j} q_t(e) d{e} \nonumber
\end{align}
Similarly for charge bids
\begin{align}
    B_{t,j} &= \frac{1}{E} \int_{E_{j-1}}^{E_j} \frac{\partial}{\partial b_t} (c p_t - Q_t(e_{t-1} - p_t/\eta + b_t\eta)) d{e_{t-1}}\nonumber\\
    &= \frac{1}{E} \int_{E_{j-1}}^{E_j} q_t(e_{t-1} - p_t/\eta + b_t\eta) d{e_{t-1}} \eta \nonumber\\
    &\approx \frac{\eta}{E} \int_{E_{j-1}}^{E_j} q_t(e) d{e} \nonumber
\end{align}

In the special case of one segment, i.e., bids are not dependent on SoC (the current energy storage bidding model in most wholesale markets), $E_{j-1}$ is zero or the lowest allowed SoC and $E_{j}$ is the highest allowed SoC value or the energy capacity. In this case the bids are simply based on the average marginal opportunity value $\bar{q}_t$
\begin{align}
    \bar{q}_t = \frac{1}{E} \int_{0}^E q_t(e) d{e}
\end{align}
and the discharge bid is $c + \bar{q}_t/\eta$, and the charge bid is $\bar{q}_t\eta$.

\subsection{Real-time market clearing and arbitrage simulation \label{arbapp}}
We consider a simplified real-time market clearing model with a generalized multi-segment energy storage bids
\begin{subequations}
\begin{gather}
\min_{p_{t,j,s},b_{t,j,s}} \quad J_t(g_t) +  \sum_s\sum_j( C_{t,j,s}p_{t,j,s} - B_{t,j,s}b_{t,j,s}) \label{rtd_obj}\\
\text{s.t.} \quad e_{t,j,s} - e_{t-1,j,s} = b_{t,j,s}\eta - p_{t,j,s}/\eta \label{rtd_c1} \\
0 \leq e_{t,j,s} \leq E_{j,s} - E_{j-1,s} \label{rtd_c2} \\
g_t + \sum_s\sum_jp_{t,j,s}  = D_t + \sum_s\sum_jb_{t,j,s} : \lambda_t \label{rtd_c3}
\end{gather}
\end{subequations}
where \eqref{rtd_obj} is the objective function minimizing total bidding costs. Note that we use aggregated generator supply curve $J_t(g_t)$ and total generation $g_t$ instead of modeling the bids from each individual generator for simplicity to focus on energy storage. The second term of the objective is the discharge bids and charge bids for each energy storage $s$ and each SoC segment $j$. \eqref{rtd_c1} models the SoC evolution under single-trip efficiency $\eta$ for each SoC segment. \eqref{rtd_c2} models the upper and lower energy limit for each SoC segment, note that the minimum energy is always zero while the maximum energy for each segment is the difference between the upper and lower SoC range $E_{j,s} - E_{j-1,s}$. With the assumption of monotonic bids, \eqref{rtd_c2} guarantees storage segment transition logic in single-period dispatch. Finally, \eqref{rtd_c3} is the power balance constraint enforcing the sum of generation and storage charge/discharge equals to the total demand $D_t$ over time period $t$, the associated dual variable is thus the market clearing price $\lambda_t$.

In this paper, we assume energy storage as a price-taker. We use historical price data to simulate how the energy storage \emph{would have} been cleared in the market. In this case, we perform a Lagrangian relaxation of \eqref{rtd_c3} and move it to the objective. This decomposes the optimization into independent sub-problems for each energy storage, and for each storage, the price-taker market clearing problem is equivalent to the following price arbitrage problem
\begin{gather}
    \max_{p_{t,j},b_{t,j}} \lambda_t\sum_j(p_{t,j}-b_{t,j}) -  \sum_j( C_{t,j}p_{t,j} - B_{t,j}b_{t,j}) \label{rtd_obj1}
\end{gather}
subject to the same storage unit constraints \eqref{rtd_c1} and \eqref{rtd_c2}. Note that for this problem we omit the storage unit index $s$ as the problem formulation is the same for each storage. Hence, price-taker market clearing simulation is equivalent to arbitrage using the same bidding cost model. While we did not consider the network model in this formulation, the price-taker market clearing model is the same should we use nodal prices.

Note that the formulation in \eqref{rtd_obj1} applies to both price-taker market bidding (HA-1 and HA-10) and price response (PR-1 and PR-10). The difference is that in HA cases, storage has to decide the bids ($C_{t,j}$ and $B_{t,j}$) one hour before the market clearing period $t$, while in PR cases storage updates bids at the same time when observing the price. 
\end{document}